\documentclass[letterpaper, 10 pt, conference]{ieeeconf}
\IEEEoverridecommandlockouts
\usepackage{epsfig} %% for loading postscript figures
\usepackage{amsmath}
\usepackage{multirow}
\usepackage{tabularx}
\usepackage{color}
\usepackage{algorithmicx}
\usepackage{algpseudocode}
\usepackage{bm}
\usepackage{amsfonts}
\usepackage{graphics}
\usepackage{url}

\title{\LARGE \bf
Learning to Navigate from Simulation via Spatial and Semantic Information Synthesis with Noise Model Embedding
}

\author{Gang~Chen, Hongzhe~Yu, Wei~Dong, Xinjun~Sheng, Xiangyang Zhu and Han Ding% <-this % stops a space
\thanks{Gang~Chen, Hongzhe~Yu, Wei~Dong, Xinjun~Sheng, Xiangyang Zhu and Han Ding are with
        the State Key Laboratory of Mechanical System and Vibration,
        School of Mechanical Engineering, Shanghai Jiaotong University,
        Shanghai, 200240, China. %(Corresponding author: Wei~Dong.
        }%
        }

\begin{document}
\maketitle
\thispagestyle{empty}
\pagestyle{empty}

\begin{abstract}
While training an end-to-end navigation network in the real world is usually of high cost, simulation provides a safe and cheap environment in this training stage. However, training neural network models in simulation brings up the problem of how to effectively transfer the model from simulation to the real world (sim-to-real).
%To solve this problem, many researches focus on the development of a proper network or training method while the environment representation is often less valued.
In this work, we regard the environment representation as a crucial element in this transfer process and propose a visual information pyramid (VIP) model to systematically investigate a practical environment representation.
%In this paper, a visual information pyramid (VIP) theory is proposed to systematically investigate a practical environment representation for sim-to-real learning-based navigation.
A novel representation composed of spatial and semantic information synthesis is then established accordingly, where noise model embedding is particularly considered. %Particularly, the spatial information is presented by a noise-model-assisted depth image while the semantic information is expressed with a categorized detection image.
To explore the effectiveness of this representation, we compared the performance with representations popularly used in the literature in both simulated and real-world scenarios.
%we first extract different representations from a same dataset collected from expert operations, then feed them to the same or very similar neural networks to train the network parameters, and finally evaluate the trained neural networks in simulated and real world navigation tasks.
% To explore the effectiveness of this representation, we collected a dataset from expert operation in a simulated scenario and trained network models imported with different environment representations. %including RGB image, depth image, segmented semantic image, and our proposed information synthesis.
% The performance of these models were evaluated in another simulated scenario and a real-world scenario.
Results suggest that our environment representation stands out. %With mere one-hour-long training data collected from a very coarse simulated environment, the network model trained with our representation can successfully navigate the robot in various real scenarios with obstacles.
Furthermore, an analysis on the feature map is implemented to investigate the effectiveness through inner reaction, which could be irradiative for future researches on end-to-end navigation.

\end{abstract}

\begin{keywords}
Transfer learning, visual navigation, environment representation
\end{keywords}

%\begin{IEEEkeywords}
%visual navigation, end-to-end learning, sim-to-real
%\end{IEEEkeywords}

\section{Introduction}
%\subsubsection{Background}
The fundamental objective of mobile robot navigation is to arrive at a goal position without collision. The mobile robot is supposed to be aware of obstacles and move freely in different working scenarios.
Mathematically modeling various situations that a mobile robot may encounter is hardly possible, while end-to-end learning provides a promising data-driven solution to this high-dimension problem. % 来处理这种高维的情形
End-to-end learning maps sensor data directly to control outputs and has been proved to be promising in coping with many scenarios \cite{ConditionalImitation, LearningCrashing, DisparityDepth}.

As a data-driven approach, end-to-end learning often requires a large amount of training data. While collecting training data in the real world is usually of high cost, collecting data in simulation is much more convenient. Therefore, training the model in simulation and transfer it directly to the real world, namely sim-to-real learning, is an attractive approach. Many researchers have studied sim-to-real learning for robot manipulators \cite{Arm1, Arm2}. The working environment of a robot manipulator is usually fixed and easy to model in simulation. However, the working environments of a mobile robot are often diversified. Subtly building these working environments in simulation is hardly possible. Under such circumstances, how to transfer the network model trained in a limited number of roughly simulated environments to various real-world scenarios has to be concerned.
\begin{figure}[t]
    \centering
    \includegraphics[width=8.5cm]{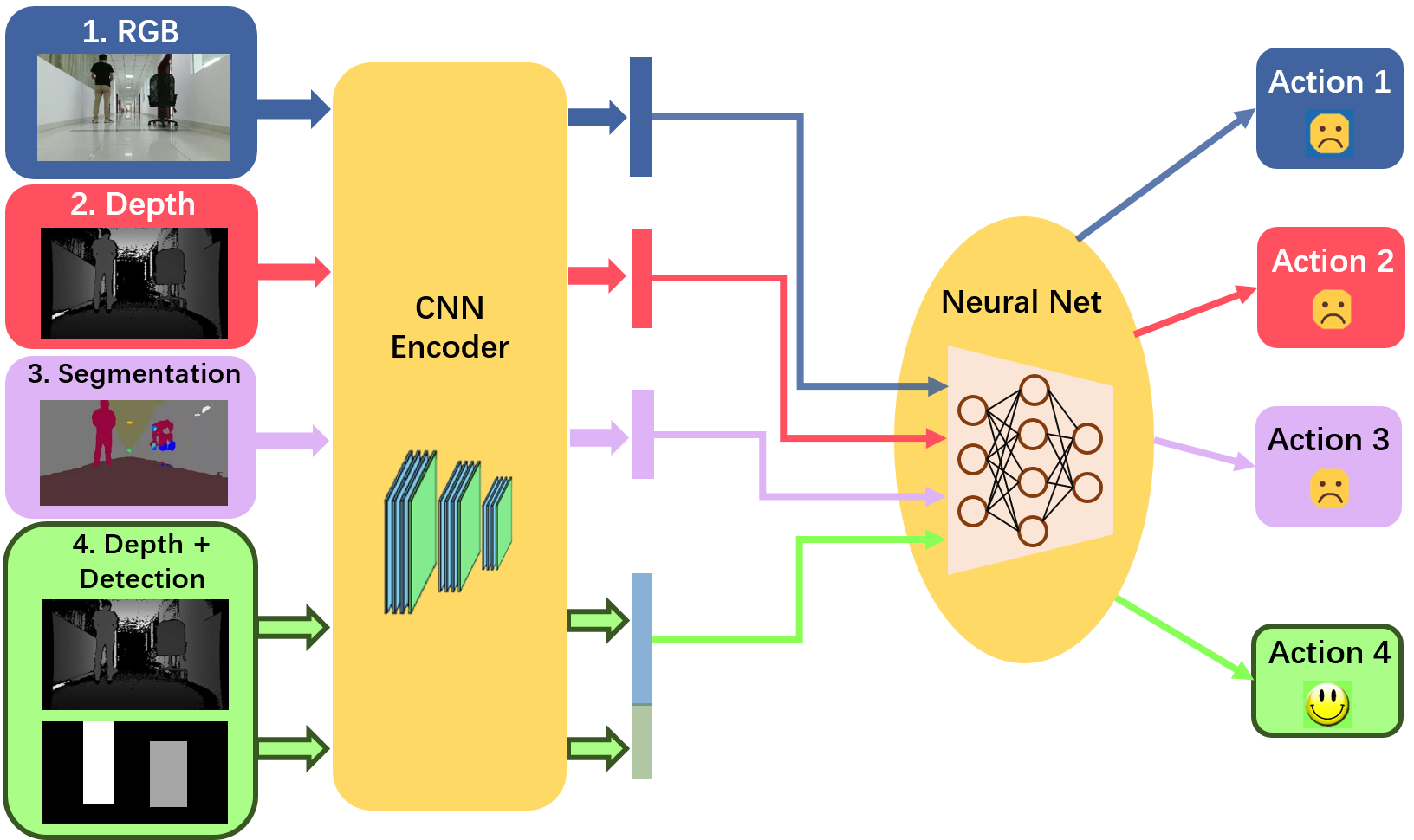}
    \caption{Method Overview: Different environment representations are compared to evaluate their influence to sim-to-real navigation tasks.}
    \label{fig:overview}
\end{figure}{}

%Therefore, high generalization ability is demanded to transfer the network model trained
%great generalization ability is demanded to train the network in a limited number of rough simulated environments and apply the generated model in various real-world environments.

To fulfill an effective transfer process, high generalization ability of the network is demanded. A crucial factor for this generalization ability is the environment representation \cite{RepresentationMattersTranferLearning}.
%The representation should contain necessary information to realize obstacle avoidance and differ little between simulation and the real world.
As vision sensor provides abundant information of the view field, state-of-the-art works use RGB image \cite{CADRL}, depth image \cite{GANSocially} and segmented semantic image \cite{Sim-Real-Driving} as the representation in end-to-end navigation. However, no systematical analysis approach is raised to compare these representations and explore a better one.
%\subsubsection{Our Approach}
Inspired by the visual information abstraction behavior of human operators, we propose the VIP model for vision-based end-to-end navigation and derive three criteria for a feasible representation in sim-to-real learning.
Accordingly, an environment representation composed of spatial and semantic information synthesis is designed. The spatial information is presented by a noise-model-embedded depth image while the semantic information is expressed with a categorized detection image.
Then a training dataset from expert operations in a coarse simulated scenario is obtained to compare the performance of this representation with others popularly used in the literature. Eight network models with different representations are trained and evaluated in two approaches.
%and network models with our representation together with all the other representations which are to be compared are trained.
%The performance of these models are evaluated in two approaches.
First in the commonly utilized approach, the models are tested in both simulated and real-world testing scenarios to get quantitative results. Furthermore, a fast and intuitive comparing approach is presented, which reveals the internal reactions of the network by constructing feature maps with the hidden convolutional layers of the network. Both ways indicates our representation behaves best, which supports our VIP model in turn.

%The spatial information is presented by a depth image acquired from an RGB-D camera, which gives the position of the obstacles. A particular noise model is utilized to fit the noise of the depth image in the real world. The semantic information is expressed by an image created from object detection results, which distinguishes obstacles in the view for social compliance. Both the depth image and the semantic are imported to our end-to-end learning network to control the mobile robot.
%弱化结论，extraordinary
%With this environment representation and the corresponding noise model, our network model trained in simulation performs excellently in real-world tests.
%achieves excellent generalization ability in vision-based sim-to-real learning for navigation tasks.
%增加Contribution!!!!
%Trained with merely one-hour-long data acquired from a simple simulated indoor environment, our network is able to drive a mobile robot in different kinds of real-world scenarios with obstacles and pedestrians, both indoor and outdoor.
%\subsubsection{Contributions}
The contributions of this work are:
\begin{itemize}
\item Proposed the VIP model and three criteria for the design of environment representation in vision-based sim-to-real navigation.
\item Designed a representation with spatial and semantic information synthesis. Noise model for the real-world sensor is particularly considered.
\item Presented a fast evaluation and analysis approach through constructing feature maps in CNN layers.
\end{itemize}

The remaining content is organized as follows: Section 2 describes the related work. Section 3 presents the design process of our representation and the learning paradigm. Section 4 gives the experiment results. The conclusion is drawn in Section 5.

%The related video of this work can be found at:
%\url{https://youtu.be/1lp417Y8Wb4}.

% \subsubsection{Organization}
% This work is organized as follow:

%In addition, this is the first work that explicitly utilizes a depth image and a semantic image to realize end-to-end navigation as far as we know.

\section{Related Work}
%\subsection{End-to-end learning}
End-to-end learning dates back to the 1980s \cite{EarliestEndtoEnd} and has been proved to be a promising approach in navigation tasks for mobile robots \cite{EarlyImageEndtoEnd}. Benefiting from the development of deep neural networks in recent years, the performance of end-to-end learning- based navigation has embraced a great improvement. The fundamental ability in navigation is obstacle avoidance. End-to-end learning networks have achieved compelling obstacle avoidance performance in many scenes, such as highway \cite{Deepdriving}, trail \cite{ForesTrails2} or corridor \cite{LearningCrashing}.
Global direction command given by a high level planner is also concerned in the literature \cite{ConditionalImitation, IntentionNet} to help robots make turns at intersections. A more complicated situation is in the environment with dynamic obstacles like pedestrians. The mobile robot must act more subtly and rapidly to avoid collision \cite{RGBandLidar, Pedestrian-Rich}.
%Bi et al. \cite{Pedestrian-Rich} use a learning from-intervention Dataset Aggregation (DAgger) algorithm based on imitation learning to deal with four situations with pedestrians. The training data is collected iteratively in the real world. Tai et al. \cite{GANSocially} presents a social compliance policy and uses generative adversarial imitation learning to learn from simulated pedestrians.
%\subsection{RGB image representation}

In these learning-based navigation works, training data is important. However, operating a mobile robot to collect training data in the real world is  inconvenient and time-consuming. Any damage to the environment or the robot itself could cause a lot of trouble. To enhance the efficiency in the data collection process, some researchers use the data acquired from the cameras mounted on a person \cite{ForesTrails, ForesTrails2} or a car \cite{DroneNet} to imitate the behaviors of a mobile robot. Another effective approach is to use sim-to-real learning. Tai et al. \cite{Sim-Real-Lidar10} adopt few sparse distance points measured by laser range finders as the network input and achieve a good sim-to-real transferability in indoor environments. In vision-based navigation, several works use the RGB image as the environment representation for sim-to-real learning networks. The RGB image can be directly fed to a single end-to-end network to get the control commands \cite{Sim-Real-Adversarial} or be divided into girds firstly to learn the best heading direction \cite{CADRL}. The result is excellent in simulation but less satisfying in the real-world tests. A special approach is to utilize two auto-encoders to generate a real RGB image from a simulated image \cite{Sim-Real-ImageGeneration}. This approach only suits a fixed number of simple scenarios since a refined mapping from simulated images to real images is quite difficult.
% We aim to design a network that suits sim-to-real learning in the environment with both static obstacles and dynamic pedestrians, which is more challenging than the environments of previous sim-to-real learning research works. To achieve the goal, spatial information and semantic information are fused to represent the environment.

%\subsection{Depth image representation}
Except for the usage of RGB image, depth image has also been adopted in sim-to-real learning for mobile robots. Depth image is easy to acquire from a stereo camera or an RGB-D camera and has been proved to be an effective environment representation when training with the data collected in the real world \cite{DepthCorridor, DisparityDepth}. Few works have tried to train the network with simulated depth images. One recent work tries this the depth image based sim-to-real learning in a pedestrian-rich scenario \cite{GANSocially}. The performance is excellent in simulation but barely satisfying in the real world due to the lack of modeling the noise in real depth images. The navigation model based on depth image in \cite{VisualSemanticSeg} behaves poorly out of the same reason.

%\subsection{Semantic image representation}
Motivated by the traditional free space searching and path planning paradigm, some works utilize a semantic image showing the free space area to form the environment representation. One implicit approach is to adopt an image segmentation network as semantic feature extraction layers and add new layers to output the control commands \cite{FCNForward}. Another explicit approach with better performance is to generate a semantic segmentation image first and then feeds it to another network to get waypoints \cite{Sim-Real-Driving} or velocity output \cite{VisualSemanticSeg, VisualRepresentations, PSPControl}. The above works behave well when the simulated training environment is  elaborate and the testing scenario is not cluttered. For a more practical sim-to-real navigation, a better environment representation is still demanded.

\section{Methods}
%To realize a good performance in sim-to-real transfer learning, a feasible environment representation is crucial.
To systematically analyze the vision-based environment representation and explore a feasible representation for sim-to-real learning, one theoretical model and three criteria are proposed in this section. Then a representation composed of spatial and semantic information synthesis is designed accordingly, in which the noise model is considered. Finally, the utilized training approach and network architecture are described.
% The detailed methods are presented as follows.
%achieve high sim-to-real performance

\subsection{Design Criteria}
%The popular end-to-end learning method regards raw sensor data, mostly RGB images, as the environment representation and inputs it to the networks. However, RGB images differ significantly from place to place, let alone the simulated environment to the real world. Thus it is difficult to transfer the model trained with RGB images from simulation to the real world. In our consideration, a feasible environment representation contains two necessary properties. Firstly, the contained information should be sufficient for obstacle avoidance and social compliance. Secondly, the difference between the simulation and the real world should be small.
%Inspired by a person's navigation, we consider the spatial information and the semantic information as the two important factors that must be included in mobile robot navigation.
% The representation of the environment is imported to the end-to-end network and is crucial in sim-to-real learning.
Consider a human operator who controls a mobile robot remotely based on a first-person-view (FPV) camera. The perception of the operator comes from an RGB image composed of basic intensity information ($I_{int}$) on each pixel and textual information ($I_{tex}$) given by the distribution of the intensity. From these detailed low-level information the human operator can abstract high-level spatial information ($I_{spa}$) and semantic information ($I_{sem}$) through his perception experience to the world and controls the robot based on these two kinds of high-level information. In addition , artificial neural networks has been proved to have the ability to abstract $I_{sem}$ from $I_{int}$ and $I_{tex}$ \cite{FCN, MaskRcnn} or the combination with $I_{spa}$ \cite{PointCloudSemantic, RGBDSegmentation}, as well as the ability to abstract $I_{spa}$ from $I_{int}$ and $I_{tex}$ \cite{RGBtoDepth, RGBtoDepthMultiScale}.
%Thus when an new operating environment with different intensity or texture comes, the operator can still control the robot well.
We define this abstraction process as visual information pyramid (VIP), as is illustrated by the pyramid in Fig. \ref{SemanticImage}.

$I_{spa}$ tells the position of all the objects in the environment. The importance of $I_{spa}$ is intuitional and has also been proved in neural science area \cite{Microstructure}. Works have been conducted to apply $I_{spa}$ in obstacle avoidance \cite{DisparityDepth, DepthCorridor, DepthFlight}.
Meanwhile, $I_{sem}$ enables the robot to distinguish the objects to perform complex behaviors \cite{Distinguishable, PedestrianIntention} or handle potential obstacles \cite{PotentialObstacle}.

%The low-level $I_{int}$ and $I_{tex}$ are the basis of $I_{spa}$ and $I_{sem}$ but matter little in navigation process.
In vision-based sim-to-real learning, an intuitive principle is to make the simulated input image in training stage as similar as possible to the real-world input image testing stage. Constructing sophisticated simulation environments to imitate real working environments is difficult and expensive. Therefore, utilizing a coarse simulation environment but a feasible environment representation that is able to keep the useful information while narrow the difference between simulation and the real world is a better solution. $I_{int}$ and $I_{tex}$ differ a lot from simulation to the real world and from place to place while they matter little in navigation. On the contrary, high-level $I_{spa}$ and $I_{sem}$ differ little but are significant in navigation.
Therefore, the first two criteria to design a feasible environment representation $\hat{E}$ for a high-performance sim-to-real navigation network are:
\begin{itemize}
\item The representation should express $I_{spa}$ and $I_{sem}$ as explicitly as possible.
\item The representation should contain little dispensable information like $I_{int}$ or $I_{tex}$.
\end{itemize}
which can be expressed as:
\begin{equation}
\hat{E} \in \left\{ E | \left\{I_{spa}, I_{sem}\right\} \sqsubset E, \left\{I_{int}, I_{tex}\right\} \not\sqsubset E \right\}
\label{criteria1and2}
\end{equation}
where the operator $\sqsubset$ denote one information is explicitly contained in an environment representation.

Furthermore, observation results are usually perfect in simulation but noisy in the real world. To further narrow the gap between simulation and the real world, noise model of the environment representation must be considered. Hence the third criterion is:
\begin{itemize}
\item A noise model $M$ for the representation $E$ satisfying the following condition can be built.
\end{itemize}

\begin{equation}
|| M(E_{sim}) - E_{real} || \to 0
\label{criteria3}
\end{equation}

\subsection{Environment Representation}
RGB image, depth image and segmented semantic image are usually used in former end-to-end navigation works. Let $E_{RGB}$, $E_{Dep.}$ and $E_{Seg.}$ denote the environment representation composed of these three images respectively, and the operator $\prec$ denote one information is not explicitly contained in a representation but can be inferred or partially inferred. Then we can get:

\begin{equation}
\left\{
\begin{array}{lr}
    \left\{I_{int}, I_{tex} \right\} \sqsubset E_{RGB}, \left\{I_{spa},I_{sem}\right\} \prec  E_{RGB} \\
    I_{spa} \sqsubset  E_{Dep.}, I_{sem} \prec  E_{Dep.} \\
    I_{sem} \sqsubset  E_{Seg.}, I_{spa} \prec  E_{Seg.}
\end{array}
\right.
\label{representations}
\end{equation}

In regard of sim-to-real learning approach, the representation of RGB image does not fit the first two criteria we proposed. It is difficult for an end-to-end navigation network to learn the high-level $I_{spa}$ and $I_{sem}$ directly from RGB image and control commands pairs. The representation of depth image explicitly contains $I_{spa}$ and can be quickly acquired by an RGB-D camera or a stereo camera, but the contained $I_{sem}$ is obscure. The representation of segmented semantic image acquired from deep learning explicitly describes $I_{sem}$ while $I_{spa}$ is only coarsely given by the layout of segmented objects in the image. Besides, segmented semantic image is usually generated by deep learning methods and the noise is unpredictable.

%Besides, it still take efforts for researchers to realize fast, stable and multi-scenario-suitable segmentation models working on an on-board computer.

%These two kinds of high-level information are significant in navigation and differ little from simulation to the real world. Meanwhile, the low-level textural or intensity information differs from place to place but affects little during the navigation process.

\begin{figure}
\centerline{\psfig{figure=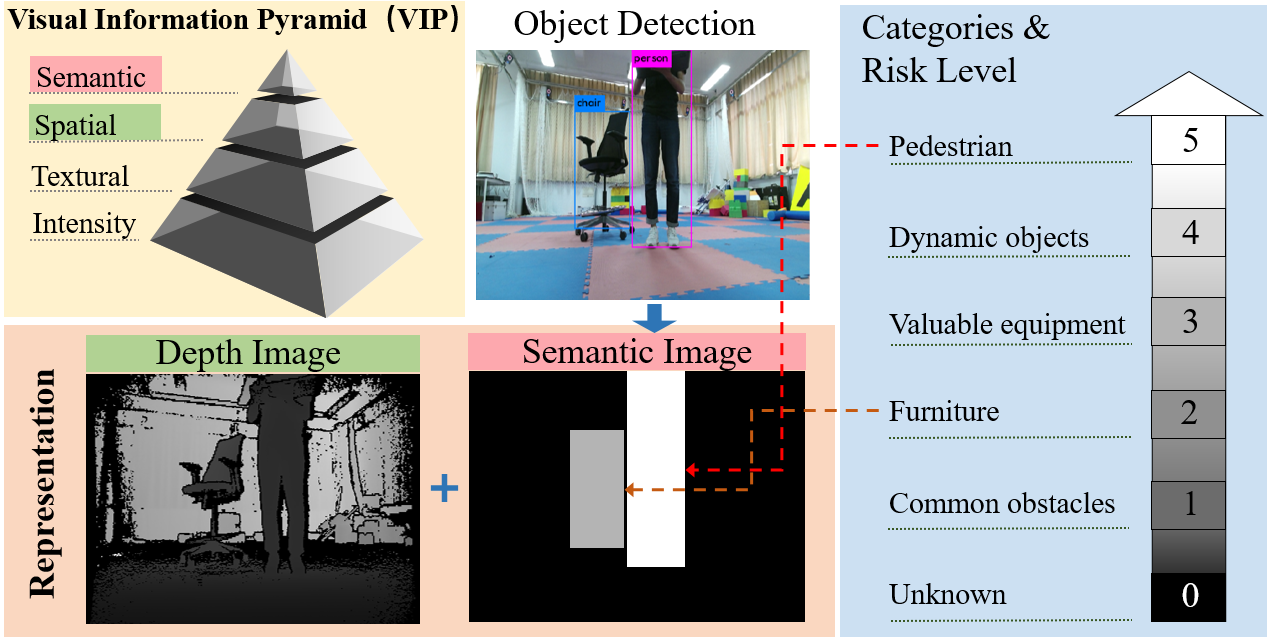,width=3.3in}}
\caption{The raised VIP theoretical model for vision based navigation and our environment representation.}
\label{SemanticImage}
\end{figure}

%Former networks that learn to navigate from simulation to the real world with RGB images are barely satisfying. The main reason is that the networks can hardly learn high-level spatial and semantic information directly from the RGB image and control commands training pairs. Mapping from intensity to control commands would fail when an environment with unseen appearance shows up.

In our environment representation, $I_{spa}$ and $I_{sem}$ are synthesized to satisfy (\ref{criteria1and2}). As is shown in Fig. \ref{SemanticImage}, a depth image considering real-world noise model is adopted to give $I_{spa}$. Additionally, a semantic image generated through the object detection results, such as the results from the Yolo V3 \cite{Yolov3}, is deployed to present $I_{sem}$.

Although object segmentation approaches could generate more elaborate semantic information containing free space area, they are not fast and accurate enough to operate with on-board computers currently.
Fast object detection results from Yolo V3 are sufficient to realize object-distinguishable obstacle avoidance. To further decrease the noise caused by false detections and increase the generalization ability, the semantic labels of the detected objects are graded into six categories according to the collision risk level. Pedestrians have the highest level which means the robot should keep a far distance and stay slow when pedestrians show up. The final semantic image is a gray-scale image that has different intensities for different categories. The higher the risk level is, the larger intensity the region is filled with. If two of the detected objects overlap, the object with a higher risk level is shown. This semantic image is named as categorized detection image to distinguish from the segmented semantic image.

Depth image obtained in the real world is pretty noisy. One type of obvious noise lies on the edges of the objects in the view, which is usually subject to a Gaussian distribution.
Our work uses the Kinect V2 RGB-D camera and the related noise model for the edges of objects has been studied in a previous work \cite{KinectModeling}. The mean of the noise is the true depth and the standard deviation can be described as:
\begin{equation}
\sigma(z)= (0.0012+0.0019(z-0.4)^2) \cdot \xi
\label{edge_noise}
\end{equation}
where $z$ is the real depth and $\xi\in[1.0, 1.2]$ is a random coefficient added to adjust extreme situations described in the previous work. Edges of objects are detected by the Canny algorithm \cite{canny}.

\begin{figure}
\centerline{\psfig{figure=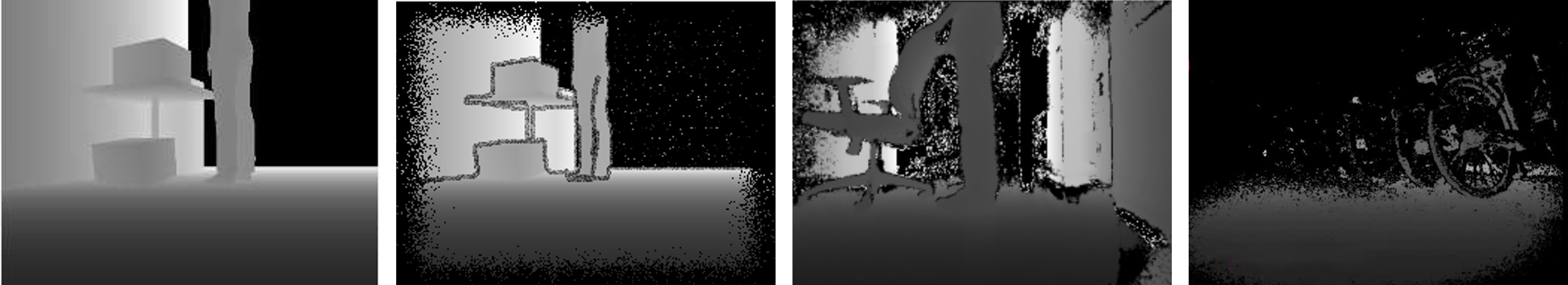,width=3.3in}}
\caption{A comparison of the simulated depth image, the simulated depth image with noise and two real-world depth images (left to right).}
\label{depth_noi}
\end{figure}

Furthermore, we found that the depth near the border of the image is often unmeasurable. The situation varies a lot in different scenarios and different light conditions hence the noise is hard to accurately model. Considering the uncertainty of this noise, a mask following a combination of Gaussian distribution and uniform distribution is added to randomly remove some values on the border of the image. The algorithm is described in Fig. \ref{edge_algorithm}, where the input ratio of the masked depth values is sampled from $0\%$ to $30\%$, $\alpha$ is 36 and $\beta$ is 24 for a depth image with $640\times480$ pixels. Finally, the salt-and-pepper noise is added randomly on the whole image. A comparison between the original simulated depth image, the noised simulated image, and two real-world depth images is shown in Fig. \ref{depth_noi}.

\begin{figure}
\centering
\framebox{\parbox{3.3in}{
\setlength{\parindent}{0.1in}
\textbf{Input:}
$\bm{P_{in}}$ (original depth image),
$\bm{r}$ (ratio of the masked depth values)

\textbf{Output:}
$\bm{P_{out}}$ (the processed depth image with noise near the edge).
\setlength{\parindent}{0.0in}

\begin{algorithmic}[1]
\State $ w \gets P_{in}.width, h \gets P_{in}.height, P_{out} \gets P_{in}$
\For{$step=0 \ to \ \frac{r \cdot w \cdot h}{2} $}
\State $(x_1, y_1) \gets (Rand(x) \mid x\sim N(0,\frac{w}{\alpha}), \, \left| x \right|\leq \frac{w}{2}, \ Rand(y) \mid y\sim U(0, h))$
\State $(x_2, y_2) \gets (Rand(x) \mid x\sim U(0,w), \ Rand(y) \mid y\sim N(0, \frac{h}{\beta}), \, \left| y \right|\leq \frac{h}{2})$
\If{$x_1<0$}
\State $x_1 \gets x_1 + w$
\EndIf
\If{$y_2<0$}
\State $y_2 \gets y_2 + h$
\EndIf
\State $P_{out}(x_1, y_1), P_{out}(x_2, y_2) \gets 0$
\EndFor
\end{algorithmic}
}}
\caption{The algorithm of adding a noise mask on the border of a depth image.}
\label{edge_algorithm}
\end{figure}

%In comparison, the networks that utilize depth image, RGB image and segmented semantic image without added noise as the environment representation are also tested in our experiments. The noise for RGB image follows the approach in \cite{ConditionalImitation}, including the change in contrast, tone and brightness and the addition of Gaussian noise, Gaussian blur, and salt-and-pepper noise. The magnitude of every transformation for each image is sampled randomly. As to the noise of segmented semantic image, we use the same segmentation deep learning model in the simulation to acquire the segmented semantic image rather than using the truth, thus the noise in simulation would be similar to the real world.
%这里可以考虑使用 E_RGB, E_dep, E_seg, E_syth表述，后面同步使用
In comparison, the network models with eight representations in four types are tested in our experiments, which are Type 1=\{RGB image ($E_{RGB}$), RGB image with noise model ($E_{RGB Noi.}$)\}, Type 2=\{depth image \cite{GANSocially} ($E_{Dep.}$), depth image with our noise model ($E_{Dep. Noi.}$)\}, Type 3=\{segmented image from FC-DenseNet \cite{FC-dense} ($E_{Seg. FC}$), segmented image from PSPNet \cite{PSPNet, PSPControl} ($E_{Seg. PSP}$)\}, and Type 4=\{our representation consists of depth image and categorized detection image ($E_{Dep. Noi. Det.}$), our representation with noise model ($E_{Dep. Det.}$)\}.
The noise added to RGB image follows the approach in \cite{ConditionalImitation}, including the change in contrast, tone and brightness and the addition of Gaussian noise, Gaussian blur, and salt-and-pepper noise.
As to the generation of the two segmented images, PSPNet is trained with ADE20k dataset \cite{ADE20k} as in \cite{PSPControl} and FC-DenseNet is trained with CamVid dataset \cite{CamVid}. In simulation, the parameters are refined with a dataset we labeled to increase the accuracy. The segmentation results are also categorized.

\subsection{Learning Approach}
In order to prove the effectiveness of our environment representation, the training data for different representations should come from the same operation process of the robot. Hence we built a dataset from expert's FPV operation in a simulated indoor environment in Gazebo \cite{Gazebo} and utilized imitation learning paradigm to train the network models with different imported representations.

There are two assumptions in imitation learning. One is that the expert performs in the right way under all encountered situations. Another is that the learning network has the input which contains all the necessary information that leads the expert to his action. The first assumption can be satisfied by carefully operating the robot to acquire good moving paths. However, learning how to recover from mistakes is also quite important \cite{DAgger}. Therefore, we randomly initialized many bad situations, such as hitting an obstacle, as the initial state to get recovering samples without bringing the operations that lead the robot to these bad situations.

The second assumption is usually fulfilled by inputting the same view that the expert had to the network. In our data collection stage, the view for the expert is an RGB image. According to our VIP model, the expert infers the spatial information and the semantic information from the RGB image and operates mainly based on these two kinds of information. Our environment representation composed of a depth image and a categorized detection image contains these two kinds of necessary information and satisfies the assumption.

To accomplish a navigation task in an environment with intersections, which is very common in many working scenarios, a global direction command also needs to be considered. Following the similar way in \cite{ConditionalAffordance}, the direction commands consist of \textit{move forward, turn left, turn right} and \textit{stop}. The expert receives the direction command at each intersection from an arrow on the screen when collecting training data, while the network takes the direction command as an input in vector form. Denote the parameters in the network $\theta$ and the expert action at a discrete time $a_j$, which includes linear velocity $v$ and angular velocity $\omega$. Assume the network can be represented by a function $F(e_j, c_j; \theta)$, where $e_j$ describes the input environment representation and $c_j$ is the global direction command. The objective of our imitation learning can be expressed as:
\begin{equation}
\operatorname{arg} \mathop{\min}_{\theta} \mathop{\sum}_{j} \operatorname{loss} \left(F(e_j, c_j; \theta), a_j \right)
\label{training_objective}
\end{equation}

\subsection{Network Architecture}
\begin{figure}
\centerline{\psfig{figure=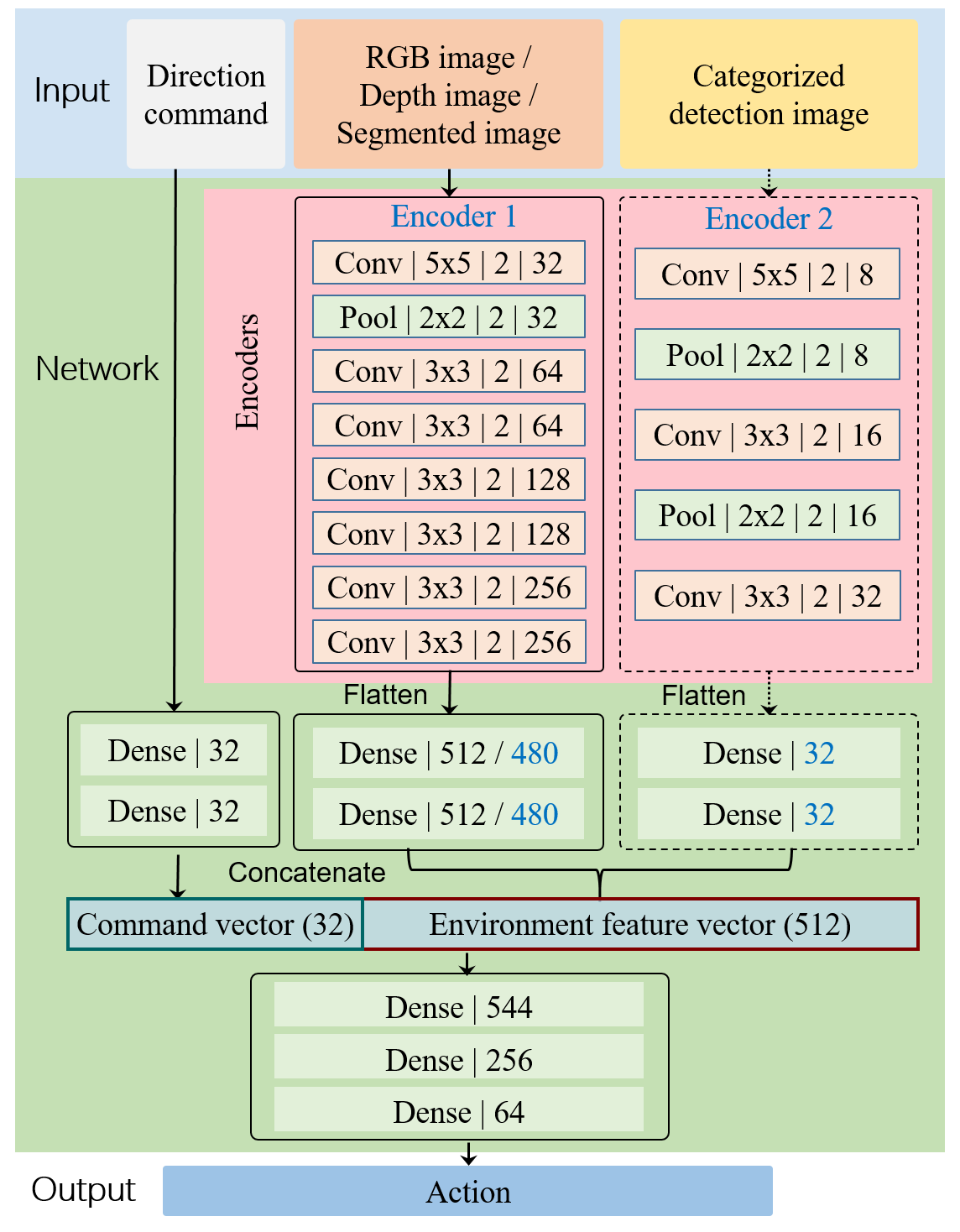,width=2.4in}}
\caption{The detailed structure of our network.}
\label{network}
\end{figure}

As has been proved in previous works on goal-directed imitation learning with image input \cite{ConditionalImitation, ConditionalAffordance},  utilizing convolutional layers to generate a vector from the input image and concatenating the vector with global direction commands is an effective network structure. The convolutional layers work as an encoder that extracts valuable features from the image. A similar modularized structure is adopted in our network. Detailed structure can be found in Fig. \ref{network}.

The input consists of environment representation and direction command. Four types of environment representations are considered for comparison as mentioned, which are based on RGB image, depth image, segmented semantic image and our representation with both depth image and categorized detection image. In the first three cases, only Encoder 1 is used to extract the features from the RGB image, the depth image or the segmented semantic image. The result is flattened and imported to two dense layers with 512 neurons to get a feature vector. After that, the feature vector is concatenated with a command vector and then connected to another three dense layers to generate the final action. In the forth case with both the depth image and the categorized detection image, two encoders are utilized. Encoder 1 stays the same except that the connected dense layers have 480 neurons each. Encoder 2 is added to extract features from the semantic image. The dense layers after Encoder 2 have only 32 neurons.

The depth image and the categorized detection image in the forth case are processed with two encoders separately. As a result, the information in the two images is not connected at the pixel-wise level. The final output can be treated as an overlay of the influence of the categorized detection image and the depth image.
%The pixel position and the depth value in the depth image give the network the spatial information of all objects in the view. Meanwhile, the intensity and pixel position in the categorized detection image give the network the semantic category of the detected objects as well as their imprecise spatial layouts.
We tried to connect the depth image and the categorized detection image at the pixel-wise level via regarding them as two channels in one image. However, the result was terrible because our categorized detection image usually contains less valuable information than the depth image, especially when there is no object detected. Thus the semantic image is valued less with a separate network.

All the images have a size of $256\times192$ pixels and the networks are light and fast to fit the real-time navigation tasks. The final output is an action vector containing linear and angular velocity control signals. Our loss function can be described as:
 \begin{equation}
\operatorname{loss}(a, a_{ref})=  {|| v-v_{ref}||}^2 + \lambda {||\omega-{\omega}_{ref}||}^2+ \sum {\gamma \theta_k^2}
\label{loss}
\end{equation}
The last component in this equation is the $L2$ regularization item. $\gamma$ is $1\times10^{-7}$ for the weights in dense layers. The range of $v$ and $\omega$ is normalized before training. $\lambda$ is a parameter to balance the effect of the error of linear and angular velocity. In practice, $\lambda = 1$ works fine.

Moreover, a 50\% dropout is applied after the convolutional layers and the first dense layer after concatenation. The ReLU nonlinearities are used for all hidden layers. The models were trained by Adam solver\cite{adam} with a mini-batch size of 40 and an initial learning rate of  $1\times10^{-4}$.

\subsection{Evaluation Approach}
The commonly utilized evaluation approach for an end-to-end navigation network model is to conduct experiments in a testing environment and assess by indicators like collision-free moving time \cite{IntentionNet} or intervention times \cite{ConditionalImitation}. For the sim-to-real paradigm, models trained in simulation should be tested on the real-world robot system to get the evaluation result. However, in the early stage of research, testing the models directly in the real world can be dangerous and time-consuming. One way is to test the models in a simulated environment different from the training environment firstly. The limitation is that the input observation is still simulated. Hence we propose a fast and intuitive approach to analyze the reaction of the network models with real observation. A typical real-world scenario with obstacles is set up and the input images are collected directly from the RGB-D camera on the robot. Then the images are fed to the network model trained in simulation and a feature map can be constructed based on the state of the middle convolutional layer. The feature map intuitively reflects the reaction of the model towards different obstacles and reveals the effectiveness of different environment representations. The details can be found in the feature map analysis part in experiments section.

\section{Experiments}
%We mainly focus on the indoor scene in our experiments.  All the original training data came from a simple simulated environment. The expert controlled the simulated mobile robot with a joy stick in the first-person view (only the RGB image can be seen). Meantime, Yolo V3 was running in the background to generate our semantic images. The RGB image, the corresponding depth image and semantic image, the direction command and the control values given by the expert were all recorded. The data was then augmented and used to train the models.
We mainly focus on indoor scenarios in our experiments. All the training data came from a very coarse simulated environment built in Gazebo \cite{Gazebo}. The results were firstly evaluated by the navigation performance both in simulation and the real world. Then our fast evaluation approach based on feature map was utilized to investigate the effectiveness through inner reaction. The following presents our system and results.

\subsection{System Setup}
\begin{figure}
\centerline{\psfig{figure=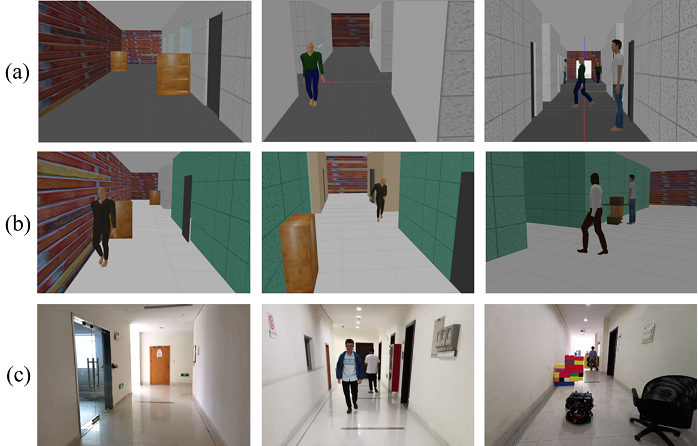,width=2.6in}}
\caption{The settings of the simulated training environment, the simulated testing environment and the real world testing environment (from Row (a) to Row (c)).}
\label{settings}
\end{figure}

\begin{figure}
\centerline{\psfig{figure=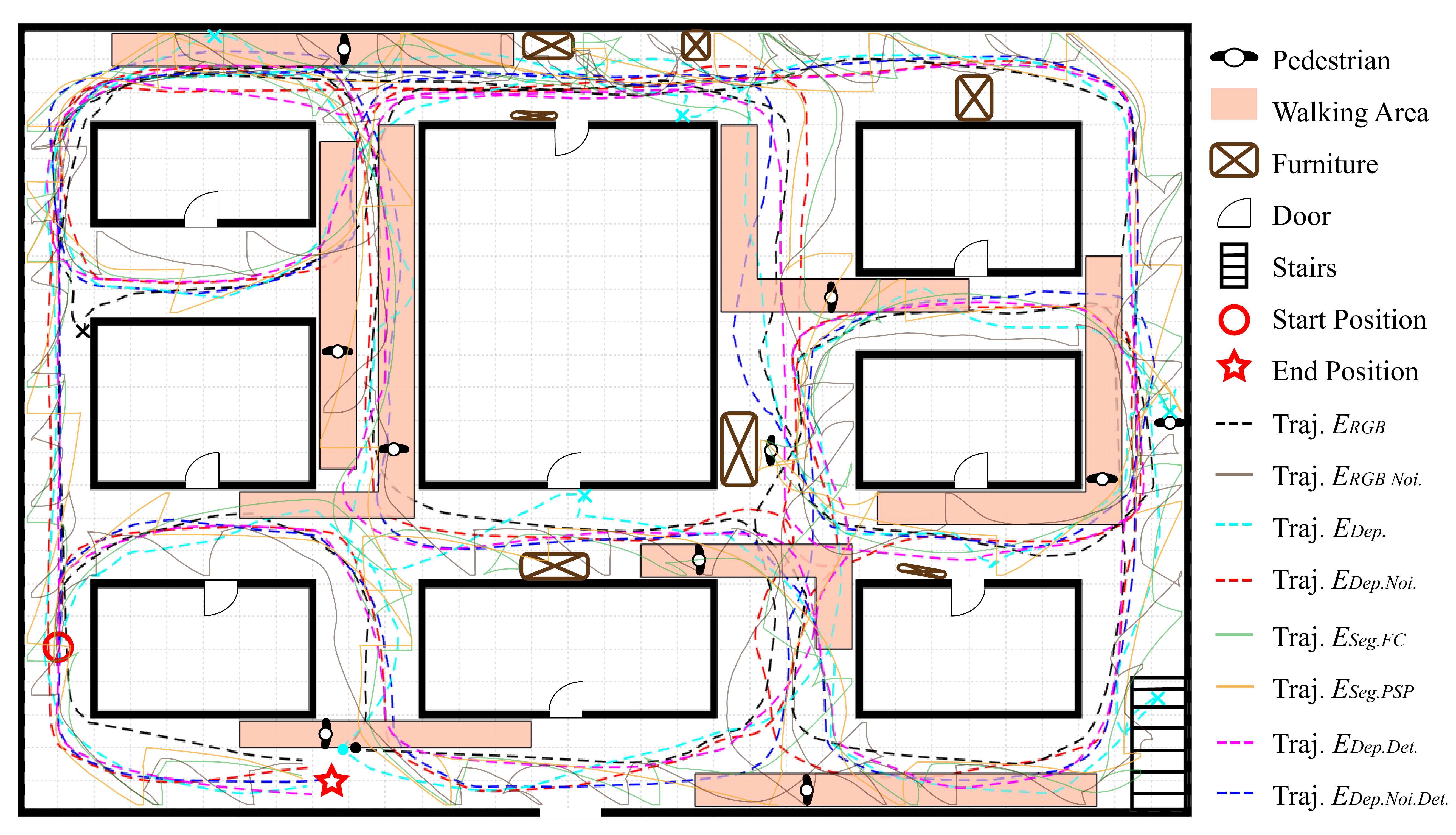,width=3.3in}}
\caption{A map of the simulated testing environment and typical trajectories of models with different environment representations.}
\label{sim_test}
\end{figure}

Two simulated indoor environments were built in Gazebo to train and test the models respectively. Compared to the training environment, the testing environment has a different building structure, and the appearances of some objects are also diverse. The settings of the simulated testing environment are shown in Fig. \ref{settings} and a map is presented in Fig. \ref{sim_test}.

In the training process, the expert controlled the simulated turtlebot with a joystick in FPV. No global route was planned and the expert followed a direction command generated randomly at each intersection. In the testing environment, the mobile followed a global route that could cover the whole map.% When a collision happens or the robot is stuck in a certain situation, we intervene and move the robot to a nearby position to continue.

\begin{figure}
\centerline{\psfig{figure=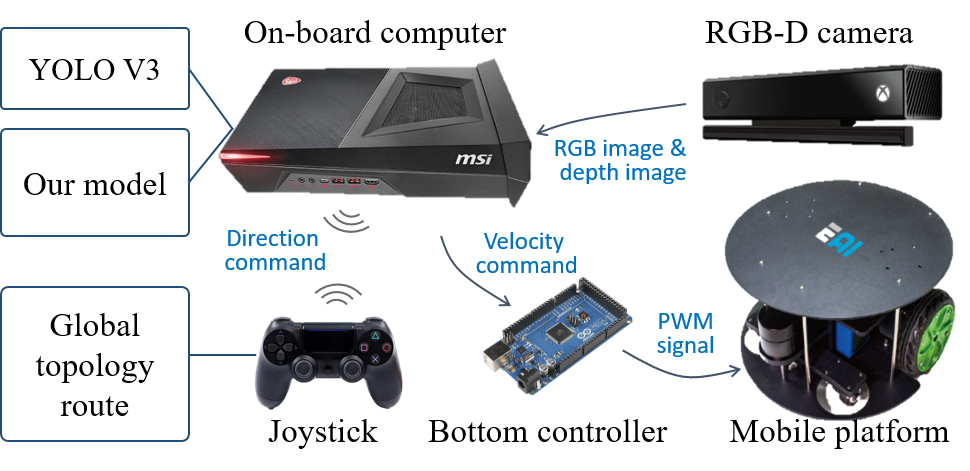,width=3.2in}}
\caption{The physical system to test the models.}
\label{physical}
\end{figure}

The physical system of the mobile robot is composed of a mobile platform, a bottom controller, an on-board computer and an RGB-D camera, as is shown in Fig. \ref{physical}. Since planning global topology route is not our focus, a joystick is simply utilized to send a direction command at each intersection. The on-board computer is mounted with an NVIDIA GTX 1060 GPU. The mobile robot moves at a speed of about 0.6 m/s.

To evaluate the models trained in simulation in the real world quantitatively, a place in our lab building was chosen as the real-world testing environment. The width of the corridors ranges from $2.0m$ to $3.2m$. The mobile robot should pass corridors with a total length of about ninety meters. The start point and the end point shared the same place. We made the corridors cluttered with some chairs and foam boards. Some voluntary pedestrians confronted, crossed or overtook the mobile robot to test its ability to react to dynamical obstacles. The volunteers had no idea of which model was running during the tests. Fig. \ref{real_test} shows a laser-scanned grid map of the real-world testing environment.

\subsection{Evaluation}
Eight models with different environment representations were trained 400 epochs each with one-hour simulated training data. We evaluated the models quantitatively in both simulated and real-world testing environments, and then analyzed the inner reaction of the networks.

In quantitative evaluations, the performance was firstly evaluated through 12 trials for each model in the simulated testing environment and 5 trials in the real-world testing environment. All our models run at a frequency of over 20 Hz in both simulation and the real-world tests. The basic obstacle avoidance ability was evaluated by the average number of intervention times. An intervention happened when the robot hit an obstacle or was stuck in a certain situation. We also evaluate the ability to react to dynamic obstacles, like pedestrians.

Previous works evaluated the ability by minimum distance to the pedestrians \cite{SociallyAware} or the successful times of avoiding hitting a pedestrian \cite{Pedestrian-Rich}. The limitation of these evaluating indicators is that they depend heavily on the behaviors of pedestrians. Pedestrians would avoid being hit and some may intentionally or occasionally walk very close to the robot. Therefore, a more objective indicator free from the behaviors of pedestrians is necessary.
Intuitively, when encountering a pedestrian, the robot decelerates, waits or turns to avoid collision. In all these actions, the linear velocity of the robot would decrease. Thus the statistical average percentage of the linear velocity decrease, compared to the situation with no pedestrian nearby, was adopted to evaluate the ability to react to moving obstacles.
The velocity was acquired by the ground truth in simulation and the motor encoders on the wheels in the real world. %%%remove
Furthermore, a score given by pedestrians after each real-world test was adopted to evaluate subjectively. Details are given below.

\begin{table}[]
\centering
\caption{Results in the simulated testing environment.}
\label{results_sim}
\begin{tabular}{cccc}
\hline
Model          & Interventions & Time (min)      & Vel. decrease \\ \hline
$E_{RGB}$           & 2.5           & 7.3             & 41.5\%            \\
$E_{RGB Noi.}$\cite{ConditionalAffordance} *      & 46.0          & 14.0              & -                \\
$E_{Dep.}$\cite{GANSocially}           & 5.6           & 10.2            & 42.2\%            \\
Our $E_{Dep. Noi.}$     & 0.6           & \textbf{5.8}    & 32.4\%            \\
$E_{Seg. FC}$  *      & 32.7          & 10.4           & -                \\
$E_{Seg. PSP}$\cite{PSPControl} *    & 26.7          & 8.1             & -                \\
Our $E_{Dep. Det.}$      & 0.9           & 6.2             & \textbf{44.8\%}   \\
Our $E_{Dep. Noi. Det.}$ & \textbf{0.2} & 6.10            & 36.1\%            \\ \hline
\end{tabular}
\end{table}

\subsubsection{Simulation Tests}
Typical trajectories in the simulated testing environment and the quantitative results are shown in Fig. \ref{sim_test} and Tab. \ref{results_sim} respectively. Representations with star indicates too many collisions happened even without dynamic obstacles, thus velocity decrease was not calculated.

The intervention of models trained with segmented images, $E_{Seg. FC}$ and $E_{Seg. PSP}$, are more than 100 times of the best model.
The model trained with non-noise depth images $E_{Dep.}$ \cite{GANSocially} behaves terribly as well, so does the model trained with non-noise RGB images $E_{RGB}$. When the augmented noisy data is added, the $E_{RGB Noi.}$ model behaves worse and collides even over 40 times in a single trial. Changes on the percentage of augmented data were also tested but helped little.
%When the percentage of augmented data increases, the behavior is worse.
On the contrary, augmented noise in the depth images helps improve the performance on obstacle avoidance significantly ($E_{Dep Noi.}$ in Tab. \ref{results_sim}). In our consideration, the augmented noise on the RGB image \cite{ConditionalAffordance}, which are designed to improve the generalization ability, never occurs in the simulated environment thus brings negative effects.
Meanwhile, depth images are much less diversified than RGB images, leading to a high possibility of overfitting. Augmented noise effectively prevents overfitting.

Compared to the models trained with only depth images, adding a categorized detection image improves the ability to avoid dynamic obstacles as expected. Moreover, the times of intervention reduces remarkably, because the semantic image raises stronger effects on the detected objects, such as furniture and pedestrians, and improves the obstacle avoidance ability. One defect is that the average time to finish one test increases slightly due to the velocity decrease. Considering the trade-off between the finishing time and the ability to safely navigate in the presence of moving obstacles, the model with our $E_{Dep. Noi. Det.}$, $E_{Dep. Det.}$ and $E_{Dep. Noi.}$ representations stand out in simulation tests.

\begin{figure}
\centerline{\psfig{figure=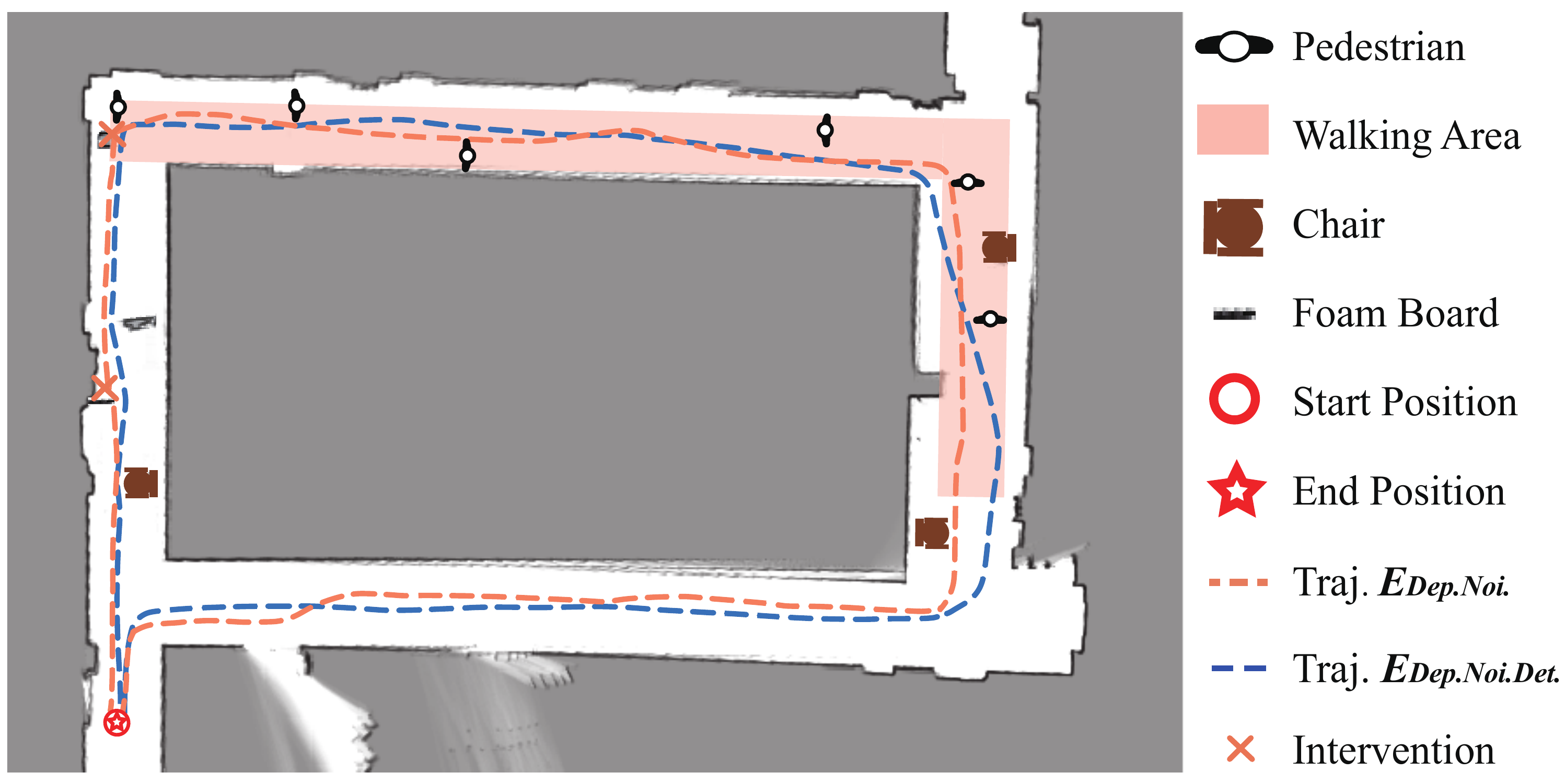,width=3.2in}}
\caption{A laser-scanned grid map of the real-world testing environment and trajectories of the two valid models.}
\label{real_test}
\end{figure}

\begin{table}[]
\centering
\caption{Results in the real-world testing environment.}
\label{results_real}
\begin{tabular}{ccccc}
\hline
Model & interventions & \begin{tabular}[c]{@{}c@{}}Time \\ (min)\end{tabular} & \begin{tabular}[c]{@{}c@{}}Vel. \\ decrease\end{tabular} & \begin{tabular}[c]{@{}c@{}}Score  \\ (0-5)\end{tabular} \\ \hline
ours $E_{Dep. Noi.}$ & 3.0 & \textbf{2.8} & 7.2\% & 3.3 \\
ours $E_{Dep. Noi. Det.}$ & \textbf{0.8} & 3.0 & \textbf{12.2\%} & \textbf{4.0} \\
Others                    &  $>$20.0 & - & - & - \\ \hline
\end{tabular}
\end{table}

\subsubsection{Real-world Tests}
Real-world tests are  much more challenging. The mobile robot moved almost randomly when using models with $E_{RGB}$ or $E_{RGB Noi.}$. When using the models with $E_{Dep.}$ or $E_{Dep. Det.}$, in which the noise in depth image is not considered, the robot just stays still or keeps steering.
Models with segmented semantic images, $E_{Seg. FC}$ and $E_{Seg. PSP}$, behave better but the number of intervention is still over 20.
Only the results of the rest two models with noise model embedding are competitive. Fig. \ref{real_test} and Tab. \ref{results_real} present the typical trajectory and quantitative results. In the ninety-meters-long real-world testing route with obstacles and pedestrians, the model with our $E_{Dep. Noi. Det.}$  shows a striking result of less than one intervention in each trial averagely. The the model with $E_{Dep. Noi.}$ also works in the real world but the performance is less excellent.
The relative performance of the two models keeps the same tendency in the simulation environment and the real-world environment.

\begin{figure}
\centerline{\psfig{figure=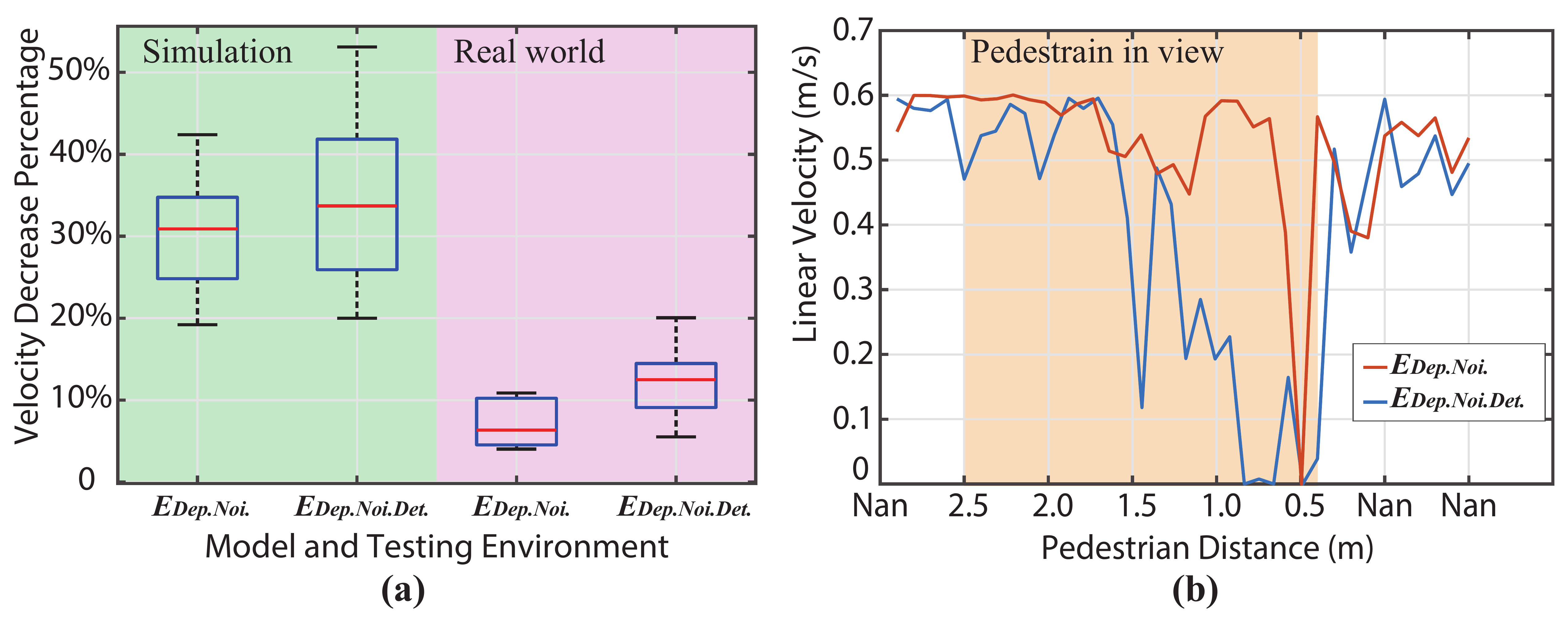,width=3.45in}}
\caption{A box plot of the velocity decrease percentage (Subplot (a)) and typical velocity command curves when the mobile robot confronts a pedestrian (Subplot (b)).}
\label{box}
\end{figure}

\begin{figure}
\centerline{\psfig{figure=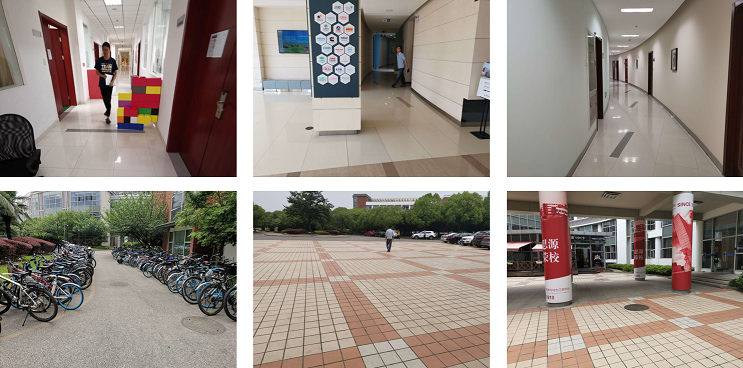,width=2.6in}}
\caption{Six testing environments for qualitative analysis.}  %% plural or sigular??
\label{morescenes}
\end{figure}

\begin{figure*}
\centerline{\psfig{figure=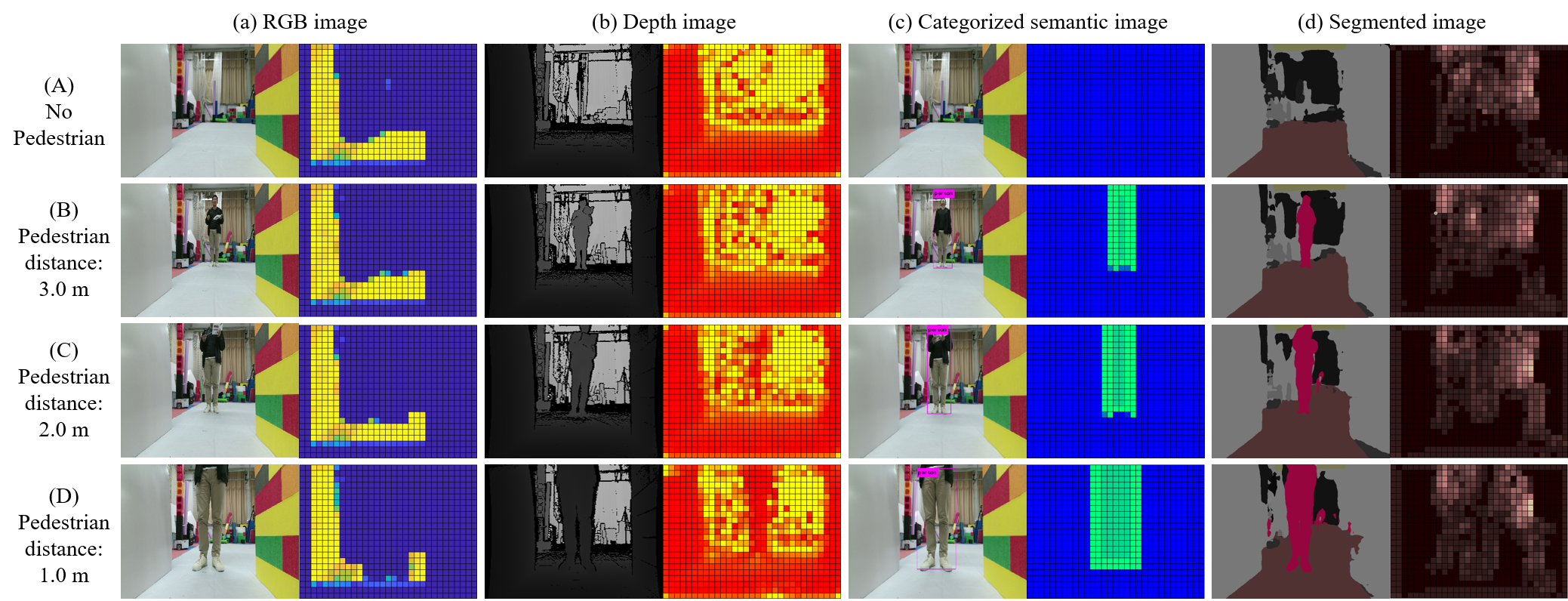,width=7.0in}}
\caption{The feature maps of the middle CNN layer for different kinds of environment representations.}
\label{inner}
\end{figure*}

% trained with both augmented depth images and semantic images
In the evaluation of the ability of avoiding moving obstacles, the model with $E_{Dep. Noi. Det.}$ behaves much better both in the velocity decrease percentage and the subjective score given by pedestrians. An illustration of the velocity decrease percentage of the two models is given by Subplot (a) in Fig. \ref{box}. Subplot (b) shows typical velocity commands given the network models. Compared to the values in simulations, the decrease percentage in the real world is less satisfying but acceptable because the behaviors of pedestrians in the real world are much more complicated. Subplot (b) presents two typical velocity command curves when the mobile robot confronts a pedestrian in the real world. The curves are aligned by the distance of the pedestrian to show the difference in the reaction distance. An obvious decrease in the output linear velocity command can be seen in both models when a pedestrian is close. The model imported with semantic image responses much earlier compared to the model without semantic image input. %One adverse effect by adding the semantic image is that the robot would occasionally miss a turn if pedestrians show up at the intersection.

Experiment in the real world shows that the model with our $E_{Dep. Noi. Det.}$, performs the best in the real world. Noise model for the depth image plays an important role. Besides, tests in a simulated environment different from the training environment could give a prior evaluation of the models before the real-world tests. We further test our best model qualitatively in six testing environments, which are shown in Fig. \ref{morescenes}. Three of the environments are indoor and the rest are outdoor.

%From the left to right and top to down order, the six environments are a cluttered corridor with pedestrians, a hall of a building, a narrow and curved corridor, a lane with many bikes, an open outdoor area and a place with pillars.
The model never met similar environments in the simulated training environment, but it is still able to avoid the obstacles. Due to the influence of sunlight, the depth image outdoors is pretty noisy while our model also performs well. The video of the experiments can be found at:
\url{https://youtu.be/ucGyuMjlgEk}.

\subsubsection{Feature Map Analysis}
To study the internal effect of the environment representations, we set up a scenario with a pedestrian and two walls with different appearances and analyzed the feature maps for different kinds of environment representations in this scenario. The analyzed network models use $E_{RGB}$, $E_{Dep. Noi.}$， $E_{Dep. Noi. Det.}$ and $E_{Seg. PSP}$. The feature map is constructed by averaging all the channels in the middle CNN layer and mapping to a gray-scale image. The middle layer is chosen because it extracts the useful features for navigation and is not too abstract to understand. The feature maps presented in Fig. \ref{inner} are recolored to have a clear view.
In the row order, the pedestrian showed up and came closer and closer to the mobile robot.

The network trained with RGB image has no obvious reaction to the pedestrian, no matter how close the pedestrian is. In the meantime, the white wall on the left is clearly shown in the feature map while the colorful wall on the right is not. The reason is that in the simulated training environment, the walls are mostly white and the people wear dark blue pants.
%Fig. \ref{rgb_more} presents one more testing situation for the RGB image trained model.
We further tested two more situations for the RGB image trained model. When the pedestrian wears a dark blue pant like the people in the simulation, some features clearly occur. When the position of the two walls is reversed, the area where the colorful wall is still has no features. This shows that the model trained with RGB image and control command pairs works mainly at the intensity level. Spatial or semantic information is hardly learned. Objects in the real world environment with different appearance raise little response.

The feature map of the model trained with depth images shows better results. The effect of walls on both sides can be seen clearly. The response to the pedestrian becomes obvious when the distance is shorter than 2 meters. When the distance is over 3 meters, the pedestrian is still clearly presented in the depth image but the network hardly responds, which means the network lowers the effect from the obstacles far-away.

The categorized semantic image is relatively simple but has the strongest response to the pedestrian. The corresponding feature map shows a rectangle representing the person, which is similar to the input semantic image. The main difference is that the intensity of the edge is higher than the inside. This is reasonable because the edge indicates the geometric layout of the pedestrian in the image, which matters in collision avoidance.

Segmented image shows reaction towards the pedestrian and the two walls. However, since the spatial information is ambiguous in the segmented image, the responses towards near obstacle and far obstacle are similar. Besides, the feature map is pretty noisy, which leads to uncertainty in obstacle avoidance.

Compared to RGB image and segmented image, depth image and categorized semantic image arouse much more obvious reactions in the real-world scenario. This accounts to the results in quantitative tests and provides a fast evaluation approach. Besides, it could also be irradiative for future researches on end-to-end navigation.

%The result further prove the effectiveness of our environment representation with spatial and semantic information synthesis.

%The semantic image and depth image in our depth-semantic-noise model are processed with two encoders separately. As a result, the information in the two images is not really connected at the pixel-wise level. The final output can be treated as an overlay of the influence of the semantic image and the depth image. The pixel position and the depth value in the depth image give the network the spatial information of all objects in the view. Meanwhile, the intensity, size and pixel position in the semantic image give the network the semantic information of the detected objects as well as their imprecise spatial layout. We had tried to input the depth image and the semantic image as two channels to one encoder to establish a pixel-wise connection. But the result was terrible because our semantic image is created from object detection results and usually contains less valuable information than the depth image, especially when there is no object detected. Thus the semantic image should be valued less with a separate network.

%The global direction command is imported to the network in a similar way as the semantic image. This separate form makes the network operate successfully even if a wrong direction command is given. For instance, when the mobile robot is passing an ``L'' shaped intersection, the mobile is able to turn to the right direction and avoid collisions even if the global direction command remains $move forward$.

\section{Conclusion}
% this work 证明了...
As the reinforcement learning-based navigation draws great attention and various learning networks are proposed, it is important to rethink the design of a proper environment representation to realize effective sim-to-real transfer learning.
This work systematically investigates the environment representation from theoretical model to evaluation approach.
%RGB image can hardly be utilized as the representation because it maps to control commands only at the intensity level while
%Bridging the gap between simulation and the real world requires high-level spatial and semantic connection from environment representation to control commands.
A representation composed of spatial and semantic information synthesis is designed accordingly. Noise model for real-world observation is particularly considered. %Both quantitative experiment results and feature map analysis prove the effectiveness of this representation.
With mere one-hour-long training data collected from a very coarse simulated environment, the network model trained with our representation can successfully navigate the robot in various real scenarios with obstacles. Feature map analysis also proves the effectiveness of this representation. In future works, we will adopt this representation in reinforcement learning-based sim-to-real navigation to improve the generalization ability. Through trial and error process in simulation, the final model is hoped to well serve the real-world navigation tasks.

%Here are two sample references: \cite{Feynman1963118,Dirac1953888}.

\bibliographystyle{IEEEtran}

\bibliography{asme}

\end{document}